\documentclass[10pt, a4paper]{article}

\usepackage{lrec-coling2024} 
\usepackage{booktabs}
\newcommand{\etdata}{WebQAmGaze}

\title{Evaluating Webcam-based Gaze Data as an \\Alternative for Human Rationale Annotations}

\name{Stephanie Brandl$^1$, Oliver Eberle$^2$, Tiago Ribeiro$^3$, \\
{\bf \large Anders Søgaard$^1$, Nora Hollenstein$^{1,4}$}}

\address{$^{1}$ University of Copenhagen, Denmark \\ 
$^{2}$ Technische Universität Berlin, Germany\\
$^{3}$ IT University of Copenhagen, Denmark\\
$^{4}$ University of Zurich, Switzerland\\
\texttt{brandl@di.ku.dk}, \texttt{oliver.eberle@tu-berlin.de}, \texttt{tiri@itu.dk},\\ \texttt{soegaard@di.ku.dk}, \texttt{nora.hollenstein@uzh.ch}}

\abstract{
Rationales in the form of manually annotated input spans usually serve as ground truth when evaluating explainability methods in NLP. They are, however, time-consuming and often biased by the annotation process. In this paper, we debate whether human gaze, in the form of webcam-based eye-tracking recordings, poses a valid alternative when evaluating importance scores.  We evaluate the additional information provided by gaze data, such as total reading times, gaze entropy, and decoding accuracy with respect to human rationale annotations. We compare WebQAmGaze, a multilingual dataset for information-seeking QA, with attention and explainability-based importance scores for 4 different multilingual Transformer-based language models (mBERT, distil-mBERT, XLMR, and XLMR-L) and 3 languages (English, Spanish, and German). Our pipeline can easily be applied to other tasks and languages. 
Our findings suggest that gaze data offers valuable linguistic insights that could be leveraged to infer task difficulty and further show a comparable ranking of explainability methods to that of human rationales.
 \\ \newline  }

\begin{document}

\maketitleabstract

\section{Introduction}

In order to build reliable and trustworthy NLP applications, it is crucial to be able to explain and interpret a model's decision. These Explainable AI (XAI) approaches often rely on human-annotated rationales for evaluation \cite{deyoung-etal-2020-eraser}. They are usually expensive and biased towards annotation guidelines \cite{hansen-sogaard-2021-guideline, parmar-etal-2023-dont} or the annotators' demographic \cite{al-kuwatly-etal-2020-identifying}.
While collecting lab-recorded human gaze data is at least as expensive as collecting rationales, it provides a more intuitive annotation process as annotators can naturally solve the task while reading the text, eliminating the need for additional post-processing annotations after task completion. Recording human gaze during annotation tasks has been suggested in the past \cite{zaidan2007using, tokunaga2013annotation}, while studies in computer vision have shown promising results when including gaze into models for attribute prediction  \cite{murrugarra2017learning}. Webcam-based eye-tracking recordings, where data is collected via a standard webcam, on the other hand, are much more cost-effective and are catching up in data quality \cite{ferhat2016low}. 

In this paper, we analyse whether and to what extent webcam-based eye-tracking can pose a valid alternative to human rationales when evaluating XAI methods in NLP. In the first part of this paper, we focus on the eye-tracking dataset itself, analysing the data quality across languages, looking for indicators of task difficulty, and evaluating to what extent gold-label rationales can be decoded directly from the eye-tracking signal. In the second part, we extend our analysis to attention-based explanations as well as Layer-Wise Relevance Propagation (LRP) \cite{transformerlrp2022} and Gradient$\,\times\,$Input \cite{JMLR:v11:baehrens10a, DBLP:journals/corr/ShrikumarGSK16}. We perform this analysis for question answering (QA) in English, Spanish, and German, and 4 multilingual Transformer models.

\begin{figure*}[t]
\centering
    \includegraphics[width=\textwidth]{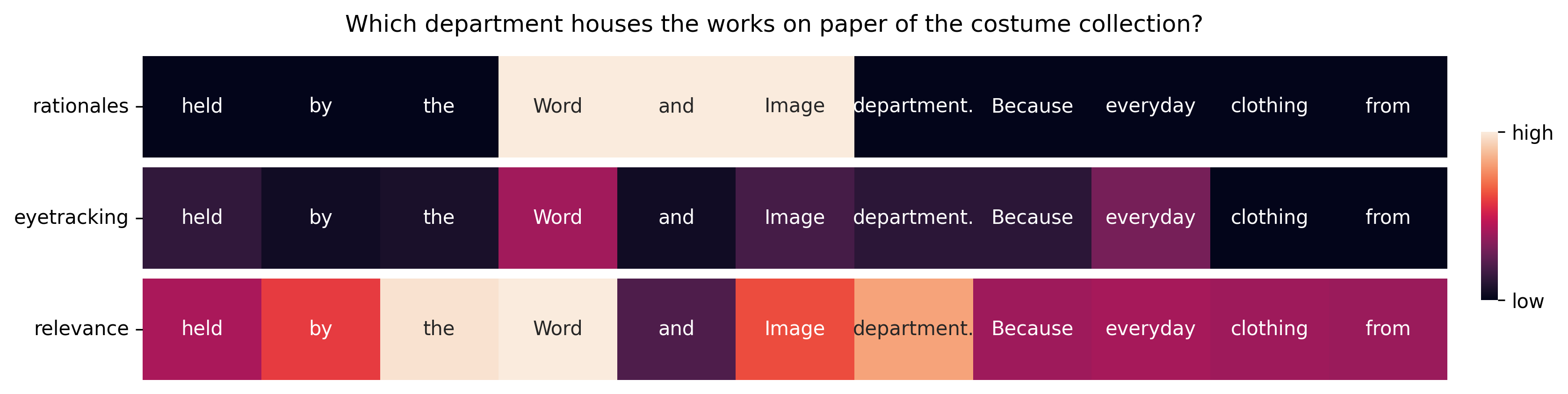}
    \caption{One sample from the WebQAmGaze corpus with ground-truth \emph{rationales}, average \emph{eye-tracking} pattern across participants and model-based \emph{relevance} scores computed with LRP based on mBERT. The correct answer is shown in the rationale (\emph{upper}). We see that both the gaze pattern and the model-based explanation scores focus on the first part of the answer more than on the second.}
    \label{fig:example}
\end{figure*}

\paragraph{Contributions.}
This work is the first analysis of webcam-based eye-tracking as an alternative for human-annotated rationales on text. We investigate (i) possible gaze-based indicators for task difficulty and (ii) factors that influence data quality in webcam-based eye-tracking. Furthermore, we (iii) fine-tune 4 multilingual Transformer models (mBERT, distilMBert, XLMR, and XLMR-L) on question answering in English, Spanish, and German to (iv) evaluate model explanations both in reference to human rationales and gaze.

We build upon the eye-tracking data and analysis of \citet{ribeiro2023webqamgaze} by computing entropy and decoding accuracy scores of the collected fixation patterns. We furthermore perform an XAI-based analysis to investigate gaze data as a possible alternative to human rationales for the evaluation of model explanations.

Results presented in Section \ref{sec:gaze_based} show that gaze provides valuable additional linguistic information that can potentially be used to infer task difficulty. Decoding accuracy based on gaze data varies across languages, resulting in promising results, especially for German. Gaze data further shows a similar ranking for explainability methods as human rationales, posing a potential alternative when evaluating XAI methods as shown in Section \ref{sec:xai}. Our code is available at: \href{https://github.com/stephaniebrandl/rationales-eyetracking-xai}{\nolinkurl{github.com/stephaniebrandl/rationales-eyetracking-xai}}.

\section{Related Work}
\paragraph{XAI for Transformers.}
Attention modules allow us to directly interpret attention tensors to understand or visualize the inner model workings \cite{bahdanau2015nmtranslation}. However, growing evidence suggests that raw attention scores may not provide a faithful explanation of the model prediction \cite{jain-wallace-2019-attention, serrano-smith-2019-attention, transformerlrp2022}. Besides the naive aggregation of raw attention weights, more elaborate explanation mechanisms have been proposed such as attention flow and attention rollout \cite{abnar-zuidema-2020-quantifying} that consider the layered model structure to assign importance scores.

Alternatively, gradient-based methods such as Gradient$\,\times\,$Input \cite{voita-etal-2019-analyzing, wu-ong-2020-explain} and integrated gradients \cite{wallace-etal-2019-allennlp} have been used to explain Transformer model predictions. However, the naive computation of model gradients in Transformers suffers from instabilities that can be mitigated via a modified layer-wise relevance propagation (LRP) scheme guided by the principle of relevance conservation \cite{transformerlrp2022, structuredxai2022}. This approach results in more faithful model explanations when compared to other Transformer explanations.

In this work, we consider a variety of methods and include both attention-based (first and last-layer attention, attention rollout) and gradient-based (Gradient$\times$Input, LRP) explanations.

\paragraph{Evaluation XAI.}
The automated quantitative evaluation of Explainable AI approaches has received growing attention \cite{ DBLP:journals/corr/abs-1911-09017, 10.5555/3463952.3463962, XAIreview2021,electronics10050593, hedstrom2023quantus}.

In evaluating explanations, it is useful to distinguish between approaches that evaluate how well a method explains the model prediction process and approaches that focus on explaining a particular ground truth. 
The former is most commonly assessed using faithfulness, sufficiency, or complexity metrics \cite{swartout}, while the latter typically involves rationale annotations or measurements to assess human alignment with the model's decision strategy \cite{Miller2019, deyoung-etal-2020-eraser}. 

We here focus on evaluating explanations based on human-annotated gold label rationales and open up the question of whether human gaze poses a valid alternative in this evaluation process.

\paragraph{Evaluation with human signals.}
Capturing the model prediction faithfully does not necessarily align with human annotations, since different task-solving strategies can emerge in models and humans \cite{rudin2019explaining, deyoung-etal-2020-eraser, atanasova-etal-2020-diagnostic}. Previous work has directly compared human and expert annotations of input data to model explanations \cite{schmidt2019,  NIPS2018_8163, deyoung-etal-2020-eraser}, and compared alignment between psychophysical signals during task-solving to model-based explanations \cite{das-etal-2016-human, klerke-etal-2016-improving,barrett-etal-2018-sequence,zhang-zhang-2019-using, hollenstein-etal-2021-multilingual}. Another line of work has analysed the alignment between human gaze and model explanations. Overall, they found a clear correlation between first-layer attention, attention flow \cite{abnar-zuidema-2020-quantifying} and gradient-based explanations with human gaze in English normal reading \cite{hollenstein-beinborn-2021-relative} as well as task-specific reading \cite{eberle-etal-2022-transformer, ikhwantri2023looking}, and also in multilingual settings \cite{morger-etal-2022-cross, brandl-hollenstein-2022-every, bensemann-etal-2022-eye}. It further has been found that higher alignment between models and gaze does not necessarily lead to higher task performance \cite{sood-etal-2020-interpreting} or  higher faithfulness \cite{eberle-etal-2022-transformer}.

\paragraph{Webcam-based eye-tracking.}
Recording human gaze via webcams enables the collection of larger datasets and has been applied in both NLP and computer vision \cite{xu2015turkergaze, papoutsaki2017searchgazer, hutt2023webcam}.
While less accurate than professional eye-tracking devices, results comparable to lab studies have been reported \cite{semmelmann_online_2018}. Results demonstrate that  well-known phenomena can be replicated from online data, but are slightly less accurate and with higher variance compared to in-lab recordings. 

We will focus our analysis on WebQAmGaze \cite{ribeiro2023webqamgaze}, a multilingual dataset for information-seeking QA. There, we compared a subset of the recorded data set with respective lab-recorded counterparts. We report Spearman correlations of greater than 0.5 for most texts. Here, we extend this and look into data quality of webcam-based eye-tracking and how this relates to decoding accuracies and evaluation in comparison to traditional XAI methods.

\begin{figure*}
    \includegraphics[width=\textwidth]{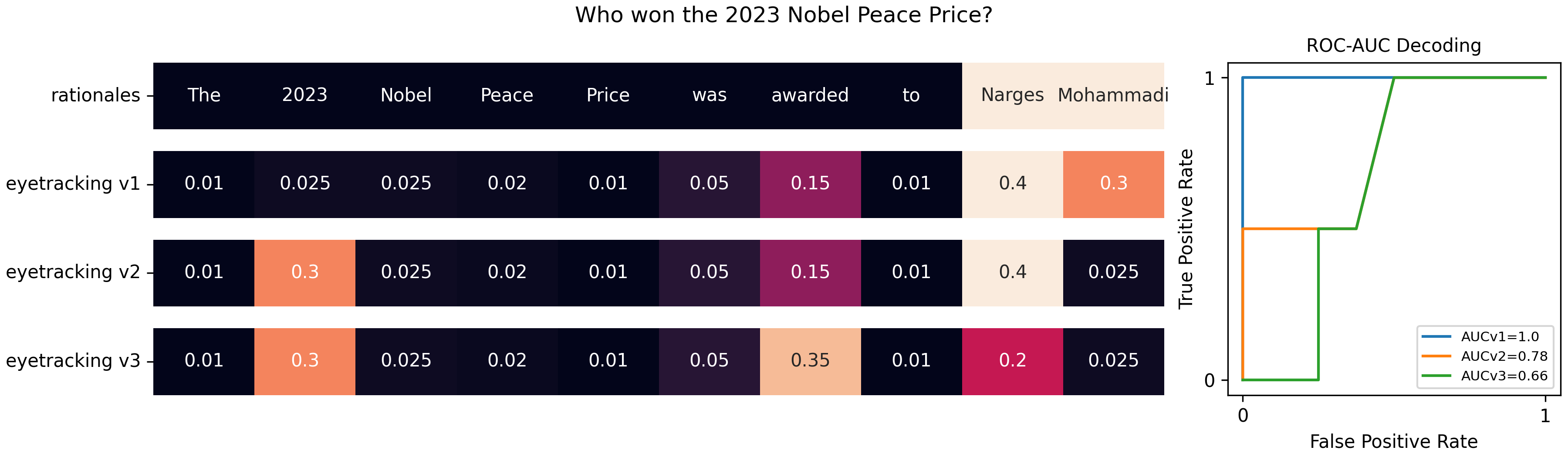}
    \caption{Toy example to visualize decoding accuracies (ROC-AUC scores) of ground-truth rationales for three different eye-tracking patterns (v1-v3). The correct ranking as in \emph{v1} leads to a perfect score of $1$. In \emph{v2} only one of the \emph{correct} tokens (\emph{Narges}) appears in the top-2 of the reading patterns which leads to a lower ROC-AUC score as shown on the right, similar for \emph{v3} where the relevant tokens only appear within the top-5. For the analysis with real gaze patterns, we only use one pattern per text in each set after averaging across participants.}
    \label{fig:roc_auc}
\end{figure*}

\section{Gaze-based Analysis} \label{sec:gaze_based}
In the following, we evaluate gaze data based on human-annotated rationales in English, Spanish, and German on a subset of the XQuAD dataset.

\subsection{Data}
\paragraph{XQUAD.}
XQuAD \cite{Artetxe:etal:2019} contains professional translations of question-answer pairs from a subset of SQuAD v1.1 \cite{rajpurkar-etal-2016-squad} into 11 languages. For each context paragraph, there is a set of questions that is annotated with the correct answer, i.e., the span of where it can be found in the text. 
The correct answers have been crowdsourced by annotators and selected based on a majority vote.
\paragraph{WebQamGaze.}
WebQamGaze \cite{ribeiro2023webqamgaze} is a multilingual webcam-based eye-tracking dataset collected with WebGazer where participants read texts from XQuAD. Participants perform two different tasks, normal reading and information-seeking, each for 4 and 5 different texts of XQuAD, respectively. In the normal reading task, each text is followed by a comprehension question. In the information-seeking (IS) task, the question is asked before showing the text but also while and after reading the respective paragraph. We focus our analysis on the information-seeking part of XQuAD for English ($N{=}126$), Spanish ($N{=}51$) and German ($N{=}19$), where $N$ represents the number of participants. Self-reported language fluency in the respective language were 4.6 - 4.9 on average per language. We show further statistics in Table \ref{tab:et}. We extract total reading times (TRT) by summing over all fixations per word and participant, i.e., how long someone looks at a specific word including regressions. We furthermore compute relative fixation duration (RFD), i.e., reading patterns, for individual participants by dividing TRT per word by the sum over all TRTs in the respective context, similar to \citet{hollenstein-beinborn-2021-relative}. Finally, we average RFD across participants. Figure \ref{fig:example} shows an example of ground-truth annotations, gaze and model explanation.
\begin{table}
\small
\resizebox{\columnwidth}{!}{
\begin{tabular}{lccccccc}
\toprule
\multicolumn{1}{c}{\textbf{Lang.}} & \multicolumn{1}{c}{\textbf{Texts}} & \multicolumn{3}{c}{\textbf{Tokens}} & \textbf{Age}  \\ 
& n & min/max& avg & answer & avg \\ 
\midrule
\textbf{EN}                        & 71                                 & 31/130       & 97       & 2.6    & 37                                    \\
\textbf{ES}                        & 42                                 & 35/131       & 96       & 2.9 & 33 \\
\textbf{DE}                        & 25                                 & 26/112       & 80       & 1.6 & 30   \\
\bottomrule
\end{tabular}
}
\caption{Statistics for the IS task in the WebQAmGaze dataset. Each row shows the number of texts, the minimum, maximum, and average number of tokens per text, followed by the average number of tokens per answer and the average age of the participants.}
\label{tab:et}
\end{table}

\subsection{Analyses}
We first carry out an in-depth analysis of \etdata. We therefore look into data quality, which varies across languages for this dataset \cite{ribeiro2023webqamgaze}. We therefore compute entropy across texts, which is known to be an indicator for task difficulty, and decoding accuracy with respect to human rationales to find out to what extent we can extract rationales from fixation patterns. This analysis aims to assess what kind of additional information the fixation patterns contain that can be beneficial for evaluating XAI.

\begin{figure}[h]
    \centering
    \includegraphics[width=0.5\textwidth]{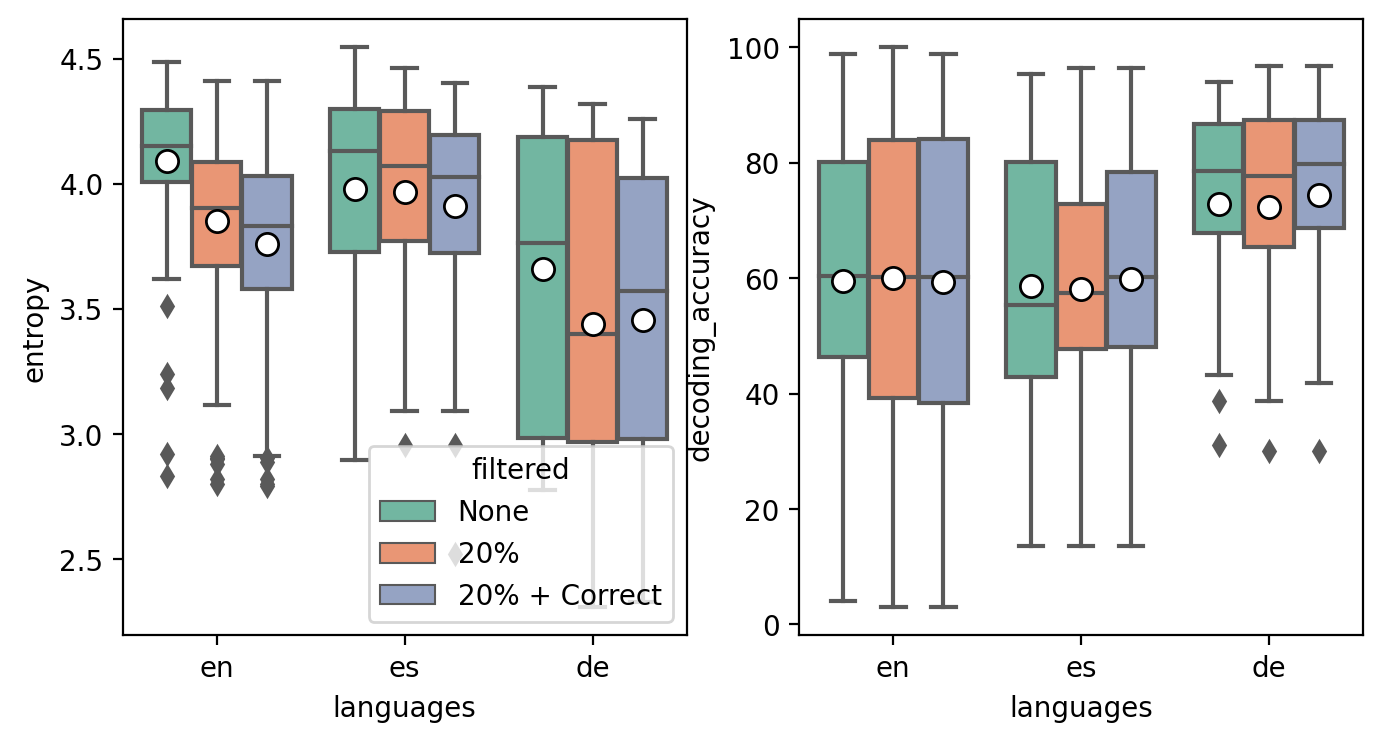}
    \caption{Entropy and decoding accuracy separated by all languages. Medians are displayed within the boxplots as a straight line whereas means are shown as white dots. Data has been filtered based on the WebGazer accuracy with a threshold of 20\% (orange) and additionally we removed wrong answers (purple).}
    \label{fig:acc_entropy_decoding}
\end{figure}

\paragraph{Data quality.}
In \citet{ribeiro2023webqamgaze}, we report WebGazer accuracies across languages as an indicator for overall data quality. The WebGazer accuracy is calculated in the calibration phase and indicates how accurately fixations are recorded on average, i.e., to what extent fixations on a particular point on the screen are detected by the software. 
We see that both median and mean across participants increase from English (23.6\%, 30.6\%) to Spanish (34.6\%, 38.4\%) and German (41.7\%, 39.0\%). This accuracy is based on individual webcams and should not be language-dependent.

\paragraph{Entropy.}
Gaze entropy, i.e., entropy calculated on fixation patterns, has been found to be an indicator for task difficulty in previous eye-tracking studies \cite{di2016gaze, wu2020eye, mejia2021gaze}. We calculate entropy across texts, results are shown in Figure \ref{fig:acc_entropy_decoding} (left), where we see a decrease in entropy of the relative fixation patterns from English to Spanish and German based on all samples, i.e., question-answer pairs. Based on the previous finding of different data quality, we filter the dataset based on WebGazer accuracy with a threshold of 20\% and keep the samples above that. We also remove wrong answers in the IS task. We find that for all languages, higher data quality and filtering out wrong answers lead to overall lower entropy values. This effect is strongest for English.

\paragraph{Error prediction.}
Based on the aforementioned literature on gaze entropy and task difficulty and workload, we look into a possible connection between error prediction in the QA task and gaze entropy. In \citet{ribeiro2023webqamgaze}, we show that TRT in WebQAmGaze differs significantly between participants who respond correctly vs.~incorrectly which here is the only available proxy for task difficulty. Our correlation analysis extends on these findings, and we observe significant negative correlation between TRT on the given text and task accuracy, i.e., the longer a person reads the text the more likely the given answer to be wrong. We find this effect to be significant in all three languages ($p<0.05$) with correlation coefficients ranging from $-0.23$ (en) to $-0.54$ (de). We further find task accuracy to correlate with entropy values ($p<0.05$) for individual samples averaged across participants for Spanish ($-0.41$) and German ($-0.49$). This suggests that higher entropy (more sparse reading patterns) correlates with a lower task accuracy, i.e., more difficult tasks, which is in line with \citet{di2016gaze, wu2020eye} who find higher entropy to be an indicator of higher workload in surgical tasks. \citet{mejia2021gaze} on the other hand, find a higher workload in driving to be connected to lower gaze entropy. All mentioned studies use 2-dimensional gaze coordinates whereas we use reading times where gaze has been allocated to specific words prior to calculating gaze entropy.

\newcolumntype{C}[1]{>{\centering\arraybackslash}m{#1}}

\begin{figure*}[h!]
    \includegraphics[width=0.95\textwidth]{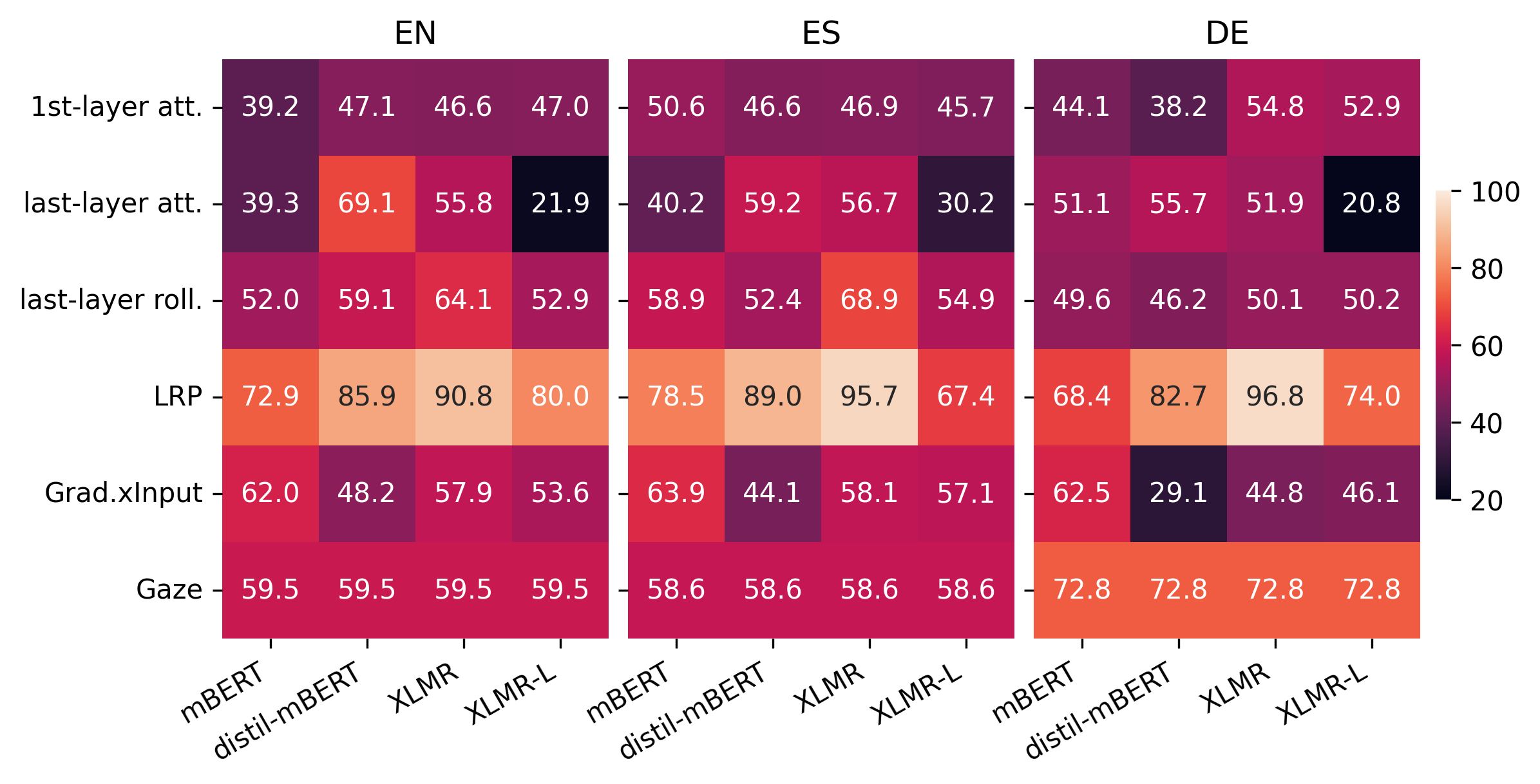}
    \caption{ROC-AUC scores for decoding rationales from attention-based and gradient-based model explanations, i.e., decoding accuracies, across all 3 languages. Results for Gaze are model-agnostic. Individual samples with an F1-scores below 50 have been filtered out per model and language.}
    \label{tab:roc_auc_xai}
\end{figure*}
\paragraph{Decoding.} \label{paragraph:gaze_decoding}

We compute decoding accuracies
to quantify how much information about the ground-truth rationale is contained in eye-tracking and model signal. This approach is related to cognitive neuroscience, where brain activity is mapped to the original stimuli (here the ground-truth rationales from XQuAD) in order to get a better understanding of human language processing \cite{huth2016natural}.

We compute ``area under the ROC curve'' (ROC-AUC) to assess the discriminatory ability of the gaze-based fixation patterns in detecting the correct rationales (true-positives) compared to the incorrectly detected tokens (false-positives). A ROC-AUC score of 0.5 indicates discrimination at chance level. Our toy example in Figure \ref{fig:roc_auc} presents three reading patterns (eye-tracking v1-3) that lead to different ROC-AUC scores as shown on the right. Here, the correct ranking is crucial, i.e., a perfect score of 1 is reached when the ranking of the top-k tokens fully agrees with the rationale tokens, here \emph{Narges} and \emph{Mohammadi}.

In Figure \ref{fig:acc_entropy_decoding} (right), we show decoding accuracies for all languages based on all samples (previously averaged across participants) and observe that rationales can indeed be decoded from gaze data with mean decoding accuracies around 60\% for English and Spanish and around 70\% for German. Here, we only observe a marginal increase after applying the same filtering with respect to data quality and task accuracy.

\begin{figure}[h]
    \centering
    \includegraphics[width=1.0\linewidth, trim={0 2.5cm 0 0} ]{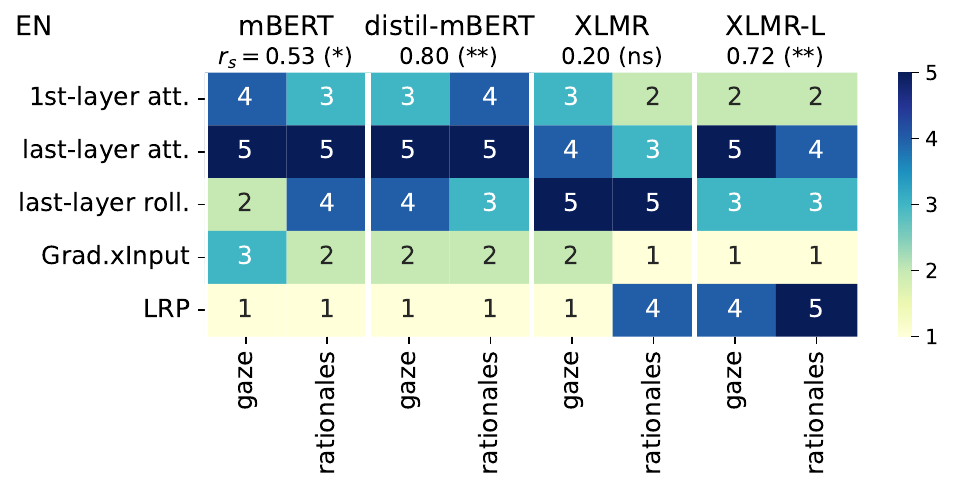}
    \includegraphics[clip, width=1.0\linewidth, trim={0 2.5cm 0 0} ]{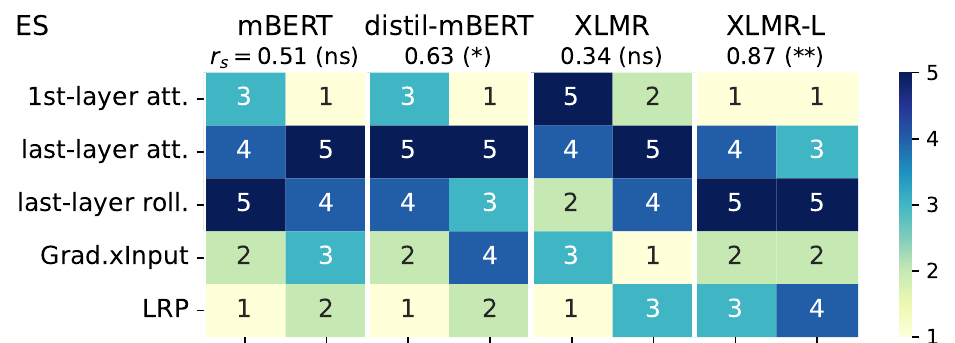}
    \includegraphics[width=1.0\linewidth, trim={0 0 0 0} ]{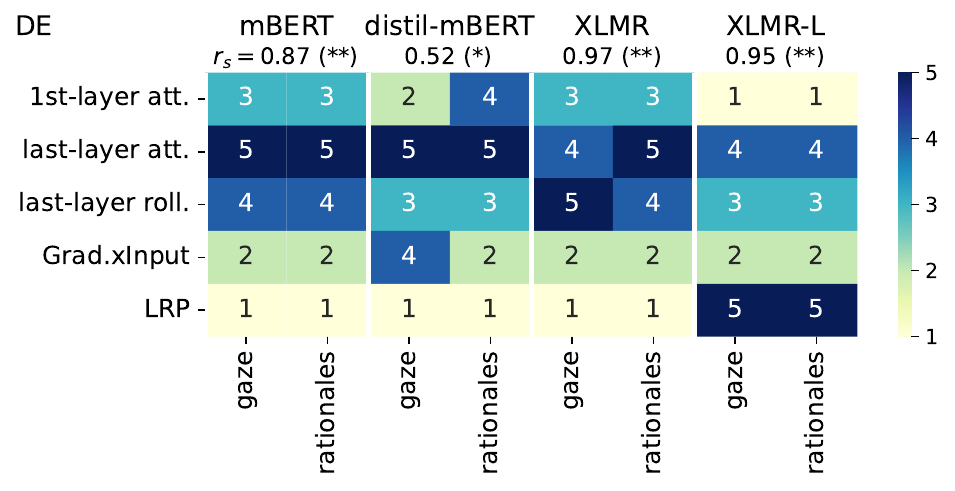}
    \caption{Comparison of gaze-based and rationale-based ranking of explanation methods for English (EN), Spanish (ES), and German (DE) -- top to bottom. Ranks 1 to 5 indicate model explanations most to least aligned with human importance scores. Spearman rank correlation $r_s$  at  $p\leq0.01$ ($^{**}$), $p\leq0.05$ ($^{*}$), or not significant (ns).
    Results are based on text samples filtered by correct human answers.}
    \label{fig:res_en_es_de}
\end{figure}
\section{XAI-based Analysis}\label{sec:xai}
We will look at the evaluation of XAI methods based on gold label rationales from two different sides.
In a first step, we extend the decoding analysis from Section \ref{sec:gaze_based} to also analyse if ground truth annotations can be decoded from model-based explanations.
In a second step, we evaluate model explanations based on ground truth rationales. This results in a ranking of XAI methods. In the following, we investigate if these rankings can similarly be obtained from gaze-based signals.

We focus our analysis on mBERT, distil-mBERT, XLMR and XLMR-L, covering a range of widely-used multilingual encoder-only models. Explanation methods based on BERT-like models have previously been shown to correlate with human gaze \cite{eberle-etal-2022-transformer, brandl-hollenstein-2022-every} as well as human rationale annotations \cite{thorn-jakobsen-etal-2023-right}.

\subsection{Fine-tuning Models}
We fine-tune 4 multilingual pre-trained language models (mBERT, distil-mBERT, XLMR, XLMR-L) individually for each of the three languages (en, es, de) on XQuAD after filtering out the language-specific text samples that have been used in WebQAmGaze. We split the remaining data into train and validation set (90/10) and use the samples from WebQAmGaze as the test set. This results in training datasets of 818-990 samples and evaluation datasets of 91-111 samples for the three languages respectively. For the fine-tuning, we use a span classification head on top of the encoder and train with AdamW with a learning rate of 2e-5, a batch size of 16, weight decay of 0.01 for 7 epochs. We fine-tune all models for 3 different seeds, evaluate all of them, and report average scores. We further filter out the samples with an F1 score below 0.5 in the QA task.

\subsection{Model Explanations}

\paragraph{Attention-based.} In order to extract model explanations, we include both attention-based and gradient-based methods. We compute averages over first-layer attention,  last-layer attention tensors  \cite{hollenstein-beinborn-2021-relative} and attention rollout \cite{abnar-zuidema-2020-quantifying}.

\paragraph{Gradient-based.}  We further compare to  `Gradient$\times$Input'  \cite{JMLR:v11:baehrens10a, DBLP:journals/corr/ShrikumarGSK16} and `Layer-wise Relevance Propagation' (LRP, \citealt{bach-plos15}).
To compute faithful LRP explanations, we apply specific propagation rules that are designed to reconstitute the conservation of relevance \cite{transformerlrp2022}. For the models considered here, this was implemented by detaching specific model components that occur in the self-attention mechanism and the normalization layers from the gradient computation. We note that this does not affect the model predictions.

\subsection{Analyses and Experiments}

\paragraph{Decoding.}

As presented in Section \ref{paragraph:gaze_decoding}, annotations could be recovered from gaze patterns with averaged ROC-AUC scores ranging from 59\% to 73\%. In Figure \ref{tab:roc_auc_xai}, we show results for the same analysis where we also decode the annotations with respect to model-based explanation for all 3 languages. We find that human rationales can be effectively decoded from explanations, in particular from LRP-based relevance with ROC-AUC scores ranging from around 67\% up to 96\% across models. 

Both first-layer and last-layer attention and rollout scores show mixed decoding abilities with ROC-AUC scores mostly below 65\% where XLMR shows the highest accuracies across all 3 languages. 

\paragraph{Ranking.}
We now turn to the question of how well human gaze patterns can be used to evaluate XAI methods in direct comparison to commonly used human rationale annotations.

To compare the agreement between annotation and explanation, we rank tokens according to their importance scores as assigned by an explanation method. We then compare the accumulated importance scores against the accumulated evidence assigned by human rationales or gaze fixations. By computing the area under the curve (AUC), we measure how well evidence from human annotations aligns with the most relevant tokens from an XAI perspective. AUC scores closer to zero indicate that model-based and gaze/rationale-based importance scores identify different tokens as most relevant for the task, AUC scores of 0.5 indicate aligned importance scores. Scores greater than 0.5 signify higher importance attribution by humans (gaze/rationales) to the most relevant tokens based on model explanation. Similar approaches have been used for the evaluation of explanation methods \cite{bach-plos15, DBLP:conf/iclr/AnconaCO018}.

The computed AUC scores are used to rank the different models, with rank 1 to 5 indicating the model explanations that are most to least aligned with human signals. The ranking across methods is shown in Figure \ref{fig:res_en_es_de} for mBERT,  distil-mBERT, XLMR, and XLMR-L. First focusing on rationales, we find that in line with previous work, gradient-based approaches rank favorably in comparison to attention-based methods (first/last-layer attention, first-layer rollout). While the respective ranking based on gaze data is less consistent, we do observe an overall comparable ranking of explanation methods, in particular, for mBERT, distil-mBERT, and XLMR. The deeper XLMR-L in comparison tends to identify first-layer-attention as the most human-aligned explanation for both gaze-based attribution and rationales. Further analysis of AUC scores suggests that for XLMR-L, AUC scores are generally lower, indicating a differently selected set of most relevant tokens.
We find that first-layer attention tends to rank favorably when compared to gaze in contrast to rationales. This effect is in line with previously observed high correlation between early-layer attention and gaze-based attention \cite{morger-etal-2022-cross, brandl-hollenstein-2022-every}. While the ranking of explanation methods can differ across models, we see that for 9/12 models (across all languages) the rankings based on gaze and rationales correlate with Spearman rank correlation scores ranging from 0.52 to 0.97 ($p\leq0.05$). This suggests that these gaze signals can be considered as an alternative to rationales for the creation of cost-effective large evaluation datasets for XAI.

\section{In-depth Analysis}
We look into different factors that potentially drive WebGazer accuracy (data quality) and decoding accuracies based on linguistic features in the text. We also show results on additional recordings for a subset of the English dataset.
\begin{figure}[h]
    \centering
    \includegraphics[width=0.5\textwidth]{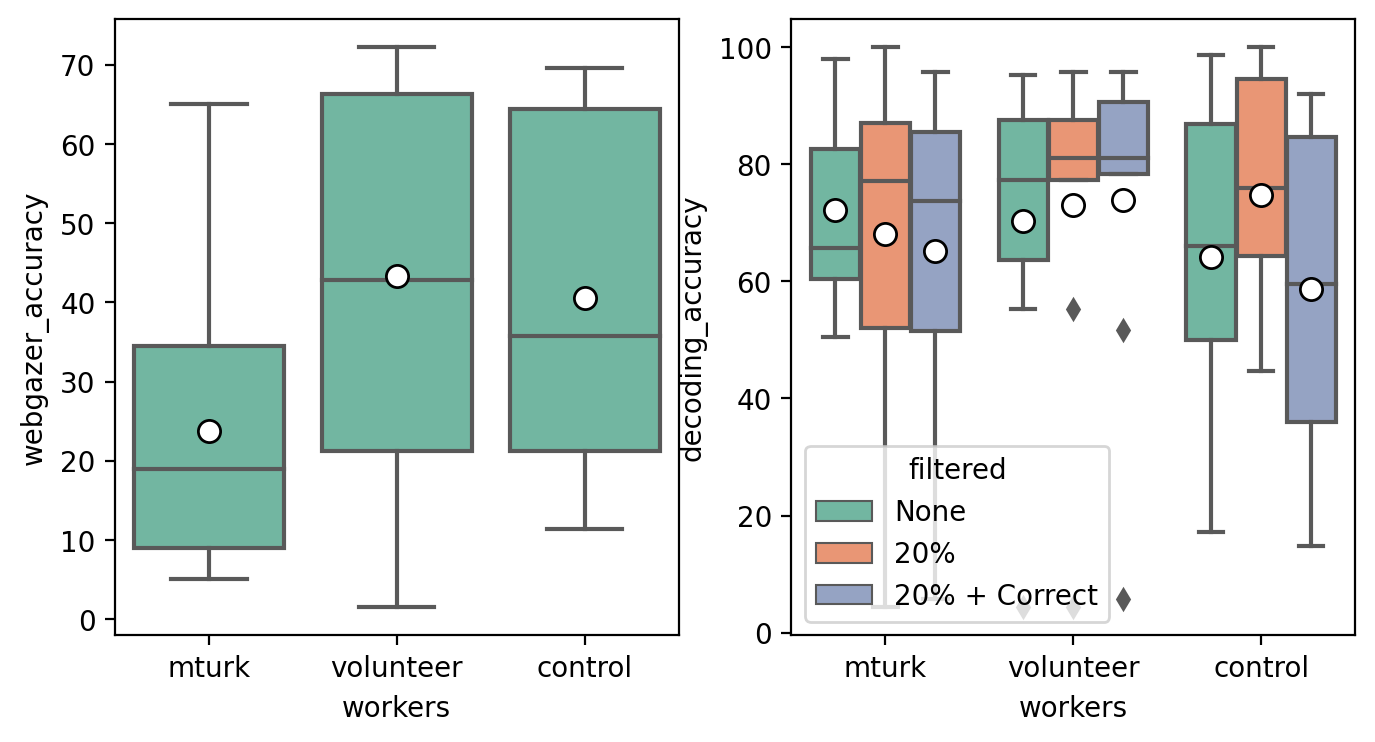}
    \caption{Comparison of WebGazer and decoding accuracies on sets 02 and 18 for different groups of workers. }
    \label{fig:volunteers}
\end{figure}
\paragraph{Control setup.}
Given the lower webcam accuracy for English, we further analyse two additional datasets that were collected for a subset of the original WebQAmGaze. The first dataset \textsc{volunteers} was recorded by 19 collaborators who were given the link online and were not paid for this study. The second dataset \textsc{control} has been recorded at the University of Copenhagen with 10 members on-site. Those participants were also not paid for this study. For the \textsc{control} dataset, we used the same laptop, a MacBook Pro 13-inch (M1, 2020), across all participants in a controlled setup where we used artificial light in a relatively dark room but no further equipment. Figure \ref{fig:volunteers} shows the results, i.e., WebGazer and decoding accuracies, for the two new datasets and for the same subset of the original dataset collected via \textsc{mturk}. We see that WebGazer accuracy still varies but is on average higher than in the same subset of the original dataset (23.8\% vs.~40.5\% and 43.4\%). Regarding decoding accuracy, the median increases for all groups when we filter based on data quality (the same holds for the mean except for \textsc{mturk}) but not the same holds for filtering out wrong answers.  We see the highest decoding accuracies for \mbox{\textsc{volunteer}} with median accuracies of 77-81\%, which is also the group with the highest WebGazer accuracy. Overall these results suggest that better webcam accuracy also leads to higher decoding accuracies.

\paragraph{Vision.}
As Figure \ref{fig:volunteers} shows, even in a controlled setup with the same lightning condition and laptop, the WebGazer accuracy varies between 11\% and 70\% with an average of 40\%. For this dataset, we collected information about participants wearing glasses during the experiment. When considering that factor, we see clear differences between people with and without glasses. For participants with glasses (4/10), the average WebGazer accuracy drops to 20\% whereas the average accuracy for people without glasses is at 54\%. We are not the first to see differences in data quality with participants wearing glasses, \citet{greenaway2021home} report that for 7 out of 9 participants, the webcam-based eye-tracker was not able to successfully mesh the participants faces due to lens reflections. Unfortunately, we do not have any information about participants' vision in the remaining datasets.

\begin{figure}[t]
    \centering
    \includegraphics[width=0.4\textwidth]{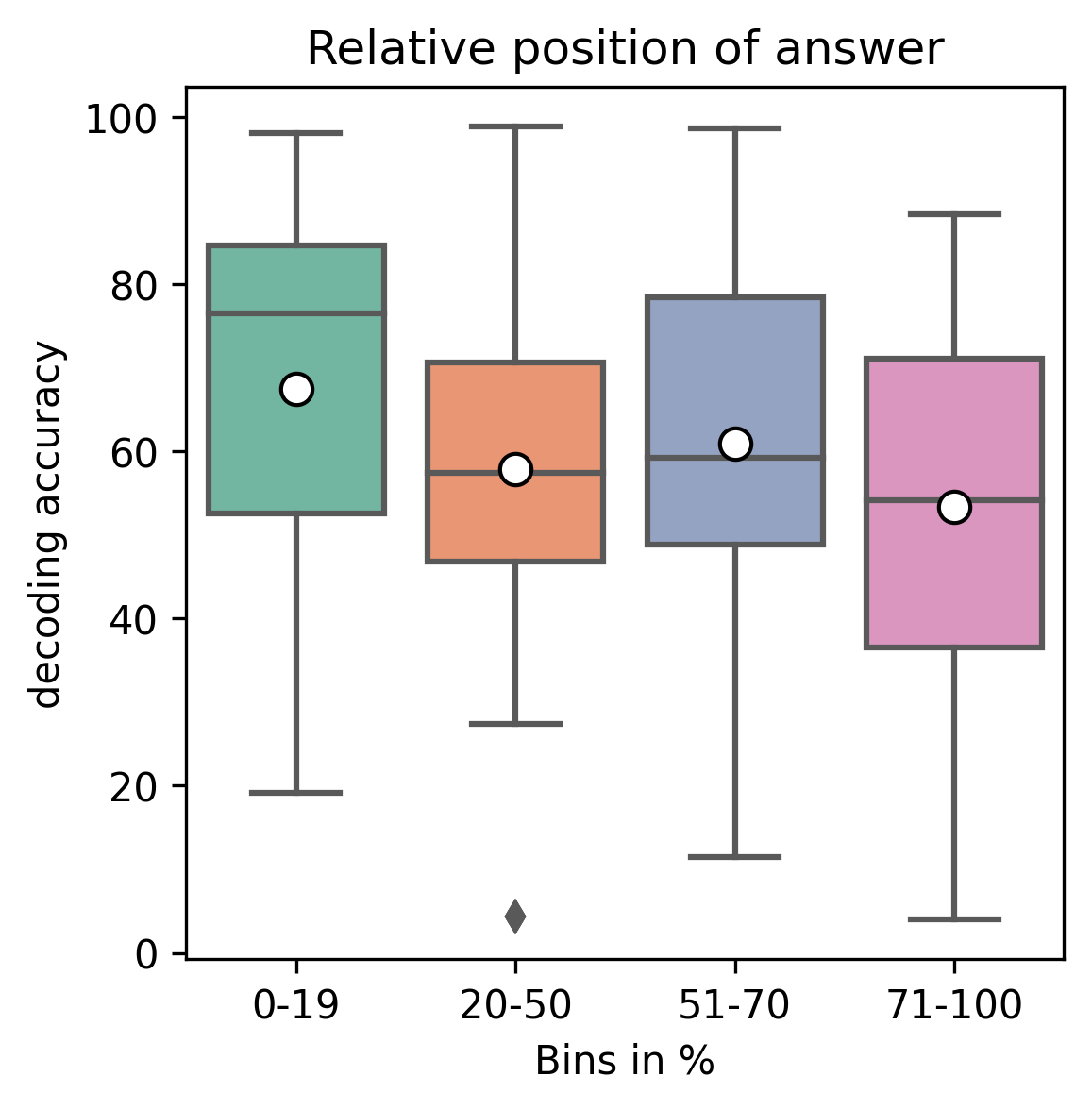}
    \caption{Decoding accuracy for the eye-tracking data with respect to ground-truth rationales based on the relative position of the answer in the text. We therefore group the dataset into 4 equally-sized bins based on where in the text the answer can be found. The x-axis shows the percentage of the upper bound of the respective bin.}
    \label{fig:position}
\end{figure}
\paragraph{Relative position of answers in text.}
We hypothesize that in the information-seeking tasks, participants stop reading after finding the correct answer which might affect decoding accuracies. We therefore split the English part of the original dataset into 4 equally sized bins based on the relative position of the answer in the text. We then compare the decoding accuracy for the ground truth rationales with respect to the eye-tracking data, similar to Figure \ref{fig:acc_entropy_decoding} (right) for individual bins and show results in Figure \ref{fig:position}. We clearly see a drop in median (and mostly also mean) accuracy the later in the text the answer can be found. 
This might be due to a more dispersed reading pattern, assuming that participants stop reading earlier, resulting in more sparse and presumably easier-to-decode reading patterns for answers located earlier in the text.

\begin{figure}[h]
    \centering
    \includegraphics[width=0.3\textwidth]{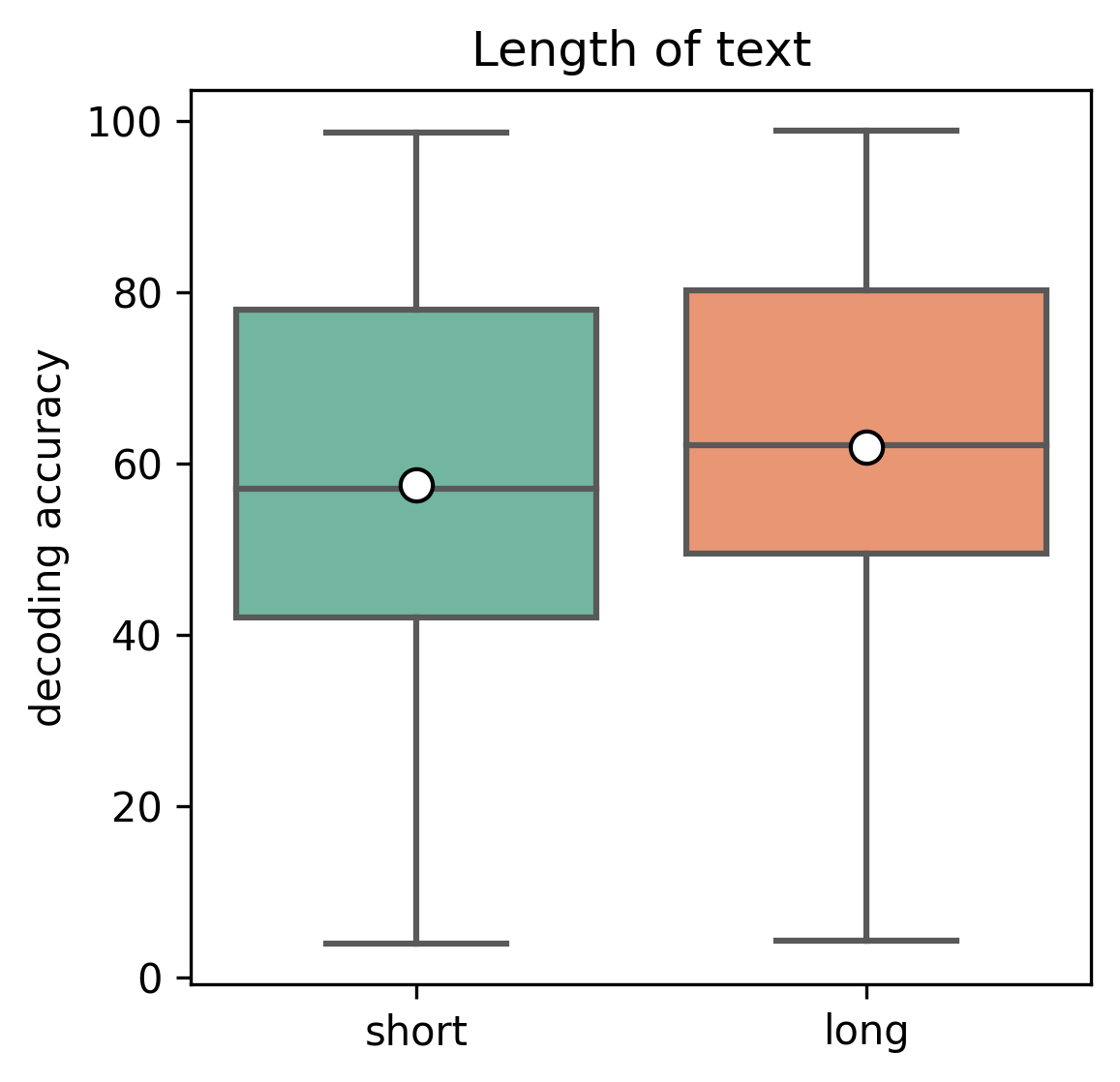}
    \caption{Results for short and long texts. English data was split into equally sized bins based on the length of the text.}
    \label{fig:len_text}
\end{figure}

\begin{figure}[t]
    \centering
    \includegraphics[width=0.5\textwidth]{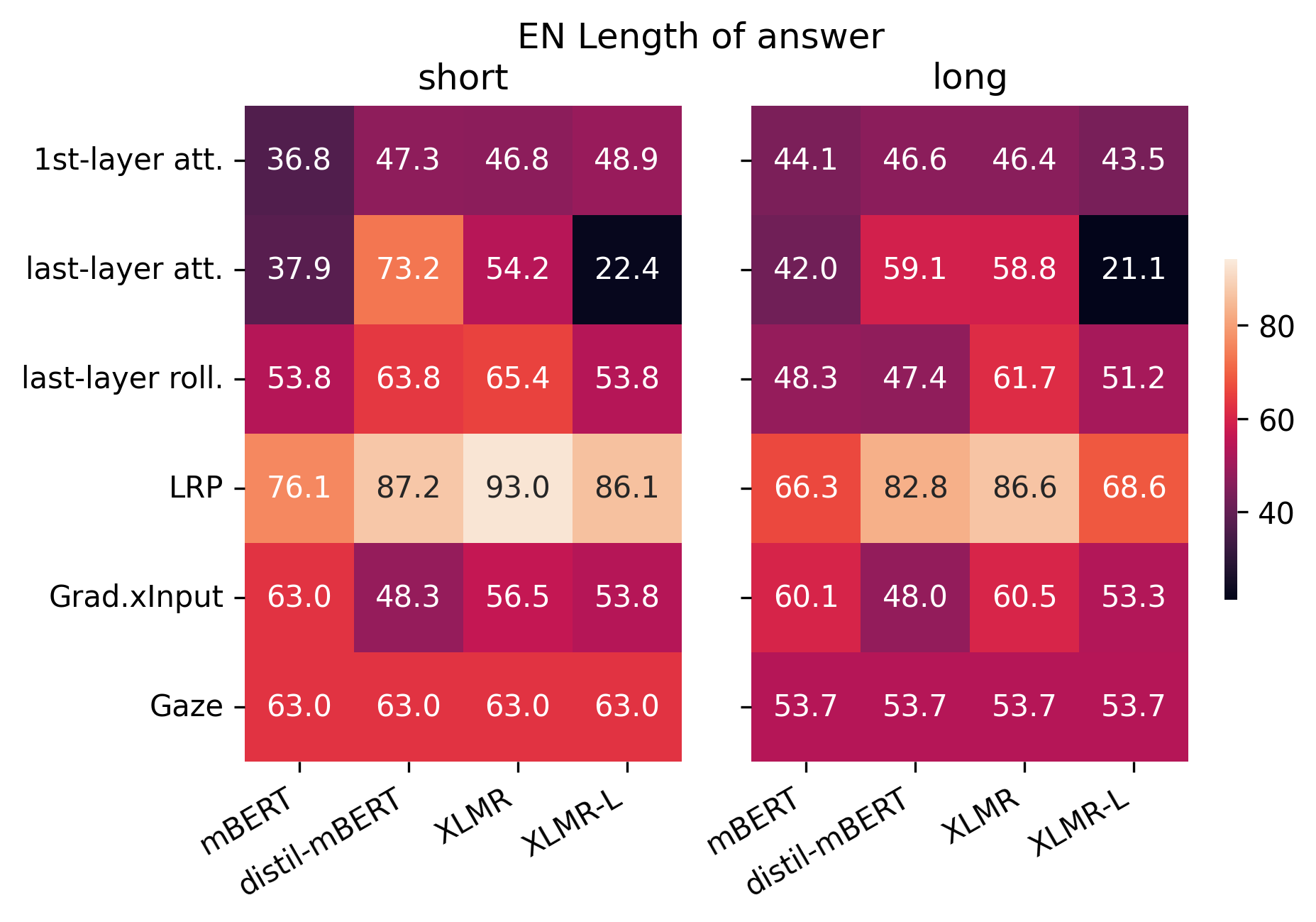}
    \caption{Results for short and long answers. English data was split into equally sized bins based on the length of the gold label answer.}
    \label{fig:len_answer}
\end{figure}
\paragraph{Length of text and answer.}
We also look into how the length of both text and answer, measured by the number of tokens, influences the decoding accuracy for eye-tracking. For both analyses, we split the English part of the original dataset into two equally sized bins. For the text length, we set the threshold at 87 tokens, i.e., the median length and show results for decoding accuracies with respect to gaze in Figure \ref{fig:len_text}. We see a slightly higher decoding accuracy for longer texts (median: 57.1\% vs.~62.1\%). Similarly, we apply the median length of all answers (2 tokens) as a threshold to split the data and look at decoding accuracies for both gaze and models in Figure \ref{fig:len_answer}. We find that models overall show slightly higher accuracies for shorter than for longer answers, while the effect for gaze is even stronger with a gap of 10\%. This effect also holds when we split the data into 4 bins.

\section{Discussion}
This work presents a first look into the possibilities of low-cost gaze data as an alternative to human rationale annotations. We have compared human gaze in an information-seeking QA task with model-based explanations in 3 languages (en, es, de) for 4 multilingual Transformer-based language models. 

We see that data quality, measured with the \mbox{WebGazer} accuracy, largely varies between recordings, even when data is collected with the same camera and lighting conditions. One reason for this might be the use of glasses, which have been shown to affect the accuracy in webcam-based eye-tracking due to lens reflections. Unfortunately, this information is typically not available and thus we recommend including it the questionnaire for future data collection. 

We use the error rate across participants as a proxy for task difficulty and look into possible indicators. We find TRT (all languages) and gaze entropy (Spanish and German) to strongly correlate with the error rate with negative coefficients. Spatial gaze entropy has been found to be an indicator for workload in surgical and driving tasks before.

Rationales and gaze provide complementary information to assess if human signals and model explanations are well-aligned. By decoding rationales from model explanations, we could clearly see that some explanations contain relevant signal to achieve high accuracies. We find that explanation methods that were found to be more faithful, in particular gradient-based explanations, are able to reach higher ROC-AUC scores than attention-based explanations. This clearly shows how the alignment of rationales and model explanations depends on the choice of appropriate XAI methods.

We see various factors that might influence decoding accuracies both for models and gaze. The relative position of the answer in the text as well as the text length and the number of tokens in the correct answer seem to potentially influence the ability to decode the gold label answer where longer texts and shorter answers lead to higher accuracies. This might be due to the fact that it takes more time, and thus more fixations are collected, to detect a shorter answer in a longer text which might lead to more accurate gaze patterns. 

We further have explored the potential use of webcam-based gaze patterns as a more accessible alternative to rationale annotations. While rankings of XAI methods that result from a comparison to (i) rationales and (ii) gaze-based attention show comparable rankings, we observe that the agreement between rankings can depend on the specific model and data.  Although evaluating and aligning models using webcam-data can currently not yet fully replace high-quality rationale annotations, we argue that they do provide useful information, in particular, when collecting rationales is not feasible. 

\section{Conclusion.}
We showed that eye-tracking, even in lower quality than lab-quality recordings, provides useful linguistic information, e.g., in the form of reading times and entropy values in English, Spanish, and German. Thus, although webcam-based eye-trackers catch up in data quality, we do not need lab-quality to benefit from the additional signals in gaze. Further research is needed to investigate entropy as a possible indicator for task difficulty in reading patterns, similar to prior work on spatial gaze patterns in workload tasks. 
This integrated approach of recording both human gaze and rationales could readily be extended to other tasks and languages, which may vary not only in linguistic but also in computational characteristics.

\section{Limitations.}
This work focuses on analysing gaze data as an alternative for human rationale annotations when evaluating explainability methods. We apply our analysis to a subset of the dataset including three Indo-European languages: English, Spanish, and German. Thus, this analysis does not cover a wide variety of languages and, furthermore, we have only a small sample size for German (19 participants). We focus on one dataset/task and on multilingual BERT-like language models. To draw more general conclusions this analysis needs to be extended to more language families, datasets, and language models.

\section{Acknowledgements}
We are particularly thankful for the volunteers at the University of Copenhagen for generously taking the time to participate in the small control study for this paper. We further thank Laura Cabello and Ilias Chalkidis for their valuable comments on the manuscript.
SB is funded by the European Union under the Grant Agreement no.~10106555, FairER. OE received funding by the German Ministry for Education and Research (under refs 01IS18056A and 01IS18025A) and BIFOLD. Views and opinions expressed are those of the author(s) only and do not necessarily reflect those of the European Union or European Research Executive Agency (REA). Neither the European Union nor REA can be held responsible for them.

\section{Bibliographical References}\label{sec:reference}

\bibliographystyle{lrec-coling2024-natbib}
\bibliography{custom}

\begin{thebibliography}{57}
\expandafter\ifx\csname natexlab\endcsname\relax\def\natexlab#1{#1}\fi

\bibitem[{Abnar and Zuidema(2020)}]{abnar-zuidema-2020-quantifying}
Samira Abnar and Willem Zuidema. 2020.
\newblock \href {https://doi.org/10.18653/v1/2020.acl-main.385} {Quantifying
  attention flow in transformers}.
\newblock In \emph{Proceedings of the 58th Annual Meeting of the Association
  for Computational Linguistics}, pages 4190--4197, Online. Association for
  Computational Linguistics.

\bibitem[{Al~Kuwatly et~al.(2020)Al~Kuwatly, Wich, and
  Groh}]{al-kuwatly-etal-2020-identifying}
Hala Al~Kuwatly, Maximilian Wich, and Georg Groh. 2020.
\newblock \href {https://doi.org/10.18653/v1/2020.alw-1.21} {Identifying and
  measuring annotator bias based on annotators{'} demographic characteristics}.
\newblock In \emph{Proceedings of the Fourth Workshop on Online Abuse and
  Harms}, pages 184--190, Online. Association for Computational Linguistics.

\bibitem[{Ali et~al.(2022)Ali, Schnake, Eberle, Montavon, M{\"{u}}ller, and
  Wolf}]{transformerlrp2022}
Ameen Ali, Thomas Schnake, Oliver Eberle, Gr{\'{e}}goire Montavon,
  Klaus{-}Robert M{\"{u}}ller, and Lior Wolf. 2022.
\newblock \href {https://proceedings.mlr.press/v162/ali22a.html} {{XAI} for
  transformers: Better explanations through conservative propagation}.
\newblock In \emph{International Conference on Machine Learning, {ICML} 2022,
  17-23 July 2022, Baltimore, Maryland, {USA}}, volume 162 of \emph{Proceedings
  of Machine Learning Research}, pages 435--451. {PMLR}.

\bibitem[{Ancona et~al.(2018)Ancona, Ceolini, {\"{O}}ztireli, and
  Gross}]{DBLP:conf/iclr/AnconaCO018}
Marco Ancona, Enea Ceolini, Cengiz {\"{O}}ztireli, and Markus Gross. 2018.
\newblock Towards better understanding of gradient-based attribution methods
  for deep neural networks.
\newblock In \emph{{ICLR} (Poster)}. OpenReview.net.

\bibitem[{Artetxe et~al.(2019)Artetxe, Ruder, and Yogatama}]{Artetxe:etal:2019}
Mikel Artetxe, Sebastian Ruder, and Dani Yogatama. 2019.
\newblock \href {http://arxiv.org/abs/1910.11856} {On the cross-lingual
  transferability of monolingual representations}.
\newblock \emph{CoRR}, abs/1910.11856.

\bibitem[{Atanasova et~al.(2020)Atanasova, Simonsen, Lioma, and
  Augenstein}]{atanasova-etal-2020-diagnostic}
Pepa Atanasova, Jakob~Grue Simonsen, Christina Lioma, and Isabelle Augenstein.
  2020.
\newblock \href {https://doi.org/10.18653/v1/2020.emnlp-main.263} {A diagnostic
  study of explainability techniques for text classification}.
\newblock In \emph{Proceedings of the 2020 Conference on Empirical Methods in
  Natural Language Processing (EMNLP)}, pages 3256--3274, Online. Association
  for Computational Linguistics.

\bibitem[{Bach et~al.(2015)Bach, Binder, Montavon, Klauschen, M{\"u}ller, and
  Samek}]{bach-plos15}
Sebastian Bach, Alexander Binder, Gr{\'e}goire Montavon, Frederick Klauschen,
  Klaus-Robert M{\"u}ller, and Wojciech Samek. 2015.
\newblock On pixel-wise explanations for non-linear classifier decisions by
  layer-wise relevance propagation.
\newblock \emph{PLoS ONE}, 10(7):e0130140.

\bibitem[{Baehrens et~al.(2010)Baehrens, Schroeter, Harmeling, Kawanabe,
  Hansen, and M{{\"u}}ller}]{JMLR:v11:baehrens10a}
David Baehrens, Timon Schroeter, Stefan Harmeling, Motoaki Kawanabe, Katja
  Hansen, and Klaus-Robert M{{\"u}}ller. 2010.
\newblock \href {http://jmlr.org/papers/v11/baehrens10a.html} {How to explain
  individual classification decisions}.
\newblock \emph{Journal of Machine Learning Research}, 11(61):1803--1831.

\bibitem[{Bahdanau et~al.(2015)Bahdanau, Cho, and
  Bengio}]{bahdanau2015nmtranslation}
Dzmitry Bahdanau, Kyunghyun Cho, and Yoshua Bengio. 2015.
\newblock Neural machine translation by jointly learning to align and
  translate.
\newblock In \emph{3rd International Conference on Learning Representations,
  {ICLR} 2015, San Diego, CA, USA, May 7-9, 2015, Conference Track
  Proceedings}.

\bibitem[{Barrett et~al.(2018)Barrett, Bingel, Hollenstein, Rei, and
  S{\o}gaard}]{barrett-etal-2018-sequence}
Maria Barrett, Joachim Bingel, Nora Hollenstein, Marek Rei, and Anders
  S{\o}gaard. 2018.
\newblock \href {https://doi.org/10.18653/v1/K18-1030} {Sequence classification
  with human attention}.
\newblock In \emph{Proceedings of the 22nd Conference on Computational Natural
  Language Learning}, pages 302--312, Brussels, Belgium. Association for
  Computational Linguistics.

\bibitem[{Bensemann et~al.(2022)Bensemann, Peng, Benavides-Prado, Chen, Tan,
  Corballis, Riddle, and Witbrock}]{bensemann-etal-2022-eye}
Joshua Bensemann, Alex Peng, Diana Benavides-Prado, Yang Chen, Neset Tan,
  Paul~Michael Corballis, Patricia Riddle, and Michael Witbrock. 2022.
\newblock \href {https://doi.org/10.18653/v1/2022.cmcl-1.9} {Eye gaze and
  self-attention: How humans and transformers attend words in sentences}.
\newblock In \emph{Proceedings of the Workshop on Cognitive Modeling and
  Computational Linguistics}, pages 75--87, Dublin, Ireland. Association for
  Computational Linguistics.

\bibitem[{Brandl and Hollenstein(2022)}]{brandl-hollenstein-2022-every}
Stephanie Brandl and Nora Hollenstein. 2022.
\newblock \href {https://aclanthology.org/2022.aacl-short.10} {Every word
  counts: A multilingual analysis of individual human alignment with model
  attention}.
\newblock In \emph{Proceedings of the 2nd Conference of the Asia-Pacific
  Chapter of the Association for Computational Linguistics and the 12th
  International Joint Conference on Natural Language Processing (Volume 2:
  Short Papers)}, pages 72--77, Online only. Association for Computational
  Linguistics.

\bibitem[{Camburu et~al.(2018)Camburu, Rockt\"{a}schel, Lukasiewicz, and
  Blunsom}]{NIPS2018_8163}
Oana-Maria Camburu, Tim Rockt\"{a}schel, Thomas Lukasiewicz, and Phil Blunsom.
  2018.
\newblock e-snli: Natural language inference with natural language
  explanations.
\newblock In S.~Bengio, H.~Wallach, H.~Larochelle, K.~Grauman, N.~Cesa-Bianchi,
  and R.~Garnett, editors, \emph{Advances in Neural Information Processing
  Systems 31}, pages 9539--9549. Curran Associates, Inc.

\bibitem[{Das et~al.(2016)Das, Agrawal, Zitnick, Parikh, and
  Batra}]{das-etal-2016-human}
Abhishek Das, Harsh Agrawal, Larry Zitnick, Devi Parikh, and Dhruv Batra. 2016.
\newblock \href {https://doi.org/10.18653/v1/D16-1092} {Human attention in
  visual question answering: Do humans and deep networks look at the same
  regions?}
\newblock In \emph{Proceedings of the 2016 Conference on Empirical Methods in
  Natural Language Processing}, pages 932--937, Austin, Texas. Association for
  Computational Linguistics.

\bibitem[{DeYoung et~al.(2020)DeYoung, Jain, Rajani, Lehman, Xiong, Socher, and
  Wallace}]{deyoung-etal-2020-eraser}
Jay DeYoung, Sarthak Jain, Nazneen~Fatema Rajani, Eric Lehman, Caiming Xiong,
  Richard Socher, and Byron~C. Wallace. 2020.
\newblock \href {https://doi.org/10.18653/v1/2020.acl-main.408} {{ERASER}: {A}
  benchmark to evaluate rationalized {NLP} models}.
\newblock In \emph{Proceedings of the 58th Annual Meeting of the Association
  for Computational Linguistics}, pages 4443--4458, Online. Association for
  Computational Linguistics.

\bibitem[{Di~Stasi et~al.(2016)Di~Stasi, Diaz-Piedra, Rieiro, Sanchez~Carrion,
  Martin~Berrido, Olivares, and Catena}]{di2016gaze}
Leandro~L Di~Stasi, Carolina Diaz-Piedra, H{\'e}ctor Rieiro, Jose~M
  Sanchez~Carrion, Mercedes Martin~Berrido, Gonzalo Olivares, and Andr{\'e}s
  Catena. 2016.
\newblock Gaze entropy reflects surgical task load.
\newblock \emph{Surgical endoscopy}, 30:5034--5043.

\bibitem[{Eberle(2022)}]{structuredxai2022}
Oliver Eberle. 2022.
\newblock \emph{Explainable structured machine learning}.
\newblock Ph.D. thesis, Technische Universität Berlin.

\bibitem[{Eberle et~al.(2022)Eberle, Brandl, Pilot, and
  S{\o}gaard}]{eberle-etal-2022-transformer}
Oliver Eberle, Stephanie Brandl, Jonas Pilot, and Anders S{\o}gaard. 2022.
\newblock \href {https://doi.org/10.18653/v1/2022.acl-long.296} {Do transformer
  models show similar attention patterns to task-specific human gaze?}
\newblock In \emph{Proceedings of the 60th Annual Meeting of the Association
  for Computational Linguistics (Volume 1: Long Papers)}, pages 4295--4309,
  Dublin, Ireland. Association for Computational Linguistics.

\bibitem[{Ferhat and Vilari{\~n}o(2016)}]{ferhat2016low}
Onur Ferhat and Fernando Vilari{\~n}o. 2016.
\newblock Low cost eye tracking: The current panorama.
\newblock \emph{Computational intelligence and neuroscience}, 2016.

\bibitem[{Greenaway et~al.(2021)Greenaway, Nasuto, Ho, and
  Hwang}]{greenaway2021home}
Anne-Marie Greenaway, Slawomir Nasuto, Aileen Ho, and Faustina Hwang. 2021.
\newblock Is home-based webcam eye-tracking with older adults living with and
  without alzheimer's disease feasible?
\newblock In \emph{Proceedings of the 23rd International ACM SIGACCESS
  Conference on Computers and Accessibility}, pages 1--3.

\bibitem[{Hansen and S{\o}gaard(2021)}]{hansen-sogaard-2021-guideline}
Victor Petr{\'e}n~Bach Hansen and Anders S{\o}gaard. 2021.
\newblock \href {https://doi.org/10.18653/v1/2021.bppf-1.2} {Guideline bias in
  {W}izard-of-{O}z dialogues}.
\newblock In \emph{Proceedings of the 1st Workshop on Benchmarking: Past,
  Present and Future}, pages 8--14, Online. Association for Computational
  Linguistics.

\bibitem[{Hedstr{\"o}m et~al.(2023)Hedstr{\"o}m, Weber, Krakowczyk, Bareeva,
  Motzkus, Samek, Lapuschkin, and H{\"o}hne}]{hedstrom2023quantus}
Anna Hedstr{\"o}m, Leander Weber, Daniel Krakowczyk, Dilyara Bareeva, Franz
  Motzkus, Wojciech Samek, Sebastian Lapuschkin, and Marina M-C H{\"o}hne.
  2023.
\newblock Quantus: An explainable ai toolkit for responsible evaluation of
  neural network explanations and beyond.
\newblock \emph{Journal of Machine Learning Research}, 24(34):1--11.

\bibitem[{Hollenstein and Beinborn(2021)}]{hollenstein-beinborn-2021-relative}
Nora Hollenstein and Lisa Beinborn. 2021.
\newblock \href {https://doi.org/10.18653/v1/2021.acl-short.19} {Relative
  importance in sentence processing}.
\newblock In \emph{Proceedings of the 59th Annual Meeting of the Association
  for Computational Linguistics and the 11th International Joint Conference on
  Natural Language Processing (Volume 2: Short Papers)}, pages 141--150,
  Online. Association for Computational Linguistics.

\bibitem[{Hollenstein et~al.(2021)Hollenstein, Pirovano, Zhang, J{\"a}ger, and
  Beinborn}]{hollenstein-etal-2021-multilingual}
Nora Hollenstein, Federico Pirovano, Ce~Zhang, Lena J{\"a}ger, and Lisa
  Beinborn. 2021.
\newblock \href {https://doi.org/10.18653/v1/2021.naacl-main.10} {Multilingual
  language models predict human reading behavior}.
\newblock In \emph{Proceedings of the 2021 Conference of the North American
  Chapter of the Association for Computational Linguistics: Human Language
  Technologies}, pages 106--123, Online. Association for Computational
  Linguistics.

\bibitem[{Huth et~al.(2016)Huth, De~Heer, Griffiths, Theunissen, and
  Gallant}]{huth2016natural}
Alexander~G Huth, Wendy~A De~Heer, Thomas~L Griffiths, Fr{\'e}d{\'e}ric~E
  Theunissen, and Jack~L Gallant. 2016.
\newblock Natural speech reveals the semantic maps that tile human cerebral
  cortex.
\newblock \emph{Nature}, 532(7600):453--458.

\bibitem[{Hutt et~al.(2023)Hutt, Wong, Papoutsaki, Baker, Gold, and
  Mills}]{hutt2023webcam}
Stephen Hutt, Aaron Wong, Alexandra Papoutsaki, Ryan~S Baker, Joshua~I Gold,
  and Caitlin Mills. 2023.
\newblock Webcam-based eye tracking to detect mind wandering and comprehension
  errors.
\newblock \emph{Behavior Research Methods}, pages 1--17.

\bibitem[{Ikhwantri et~al.(2023)Ikhwantri, Putra, Yamada, and
  Tokunaga}]{ikhwantri2023looking}
Fariz Ikhwantri, Jan Wira~Gotama Putra, Hiroaki Yamada, and Takenobu Tokunaga.
  2023.
\newblock Looking deep in the eyes: Investigating interpretation methods for
  neural models on reading tasks using human eye-movement behaviour.
\newblock \emph{Information Processing \& Management}, 60(2):103195.

\bibitem[{Jain and Wallace(2019)}]{jain-wallace-2019-attention}
Sarthak Jain and Byron~C. Wallace. 2019.
\newblock \href {https://doi.org/10.18653/v1/N19-1357} {{A}ttention is not
  {E}xplanation}.
\newblock In \emph{Proceedings of the 2019 Conference of the North {A}merican
  Chapter of the Association for Computational Linguistics: Human Language
  Technologies, Volume 1 (Long and Short Papers)}, pages 3543--3556,
  Minneapolis, Minnesota. Association for Computational Linguistics.

\bibitem[{Klerke et~al.(2016)Klerke, Goldberg, and
  S{\o}gaard}]{klerke-etal-2016-improving}
Sigrid Klerke, Yoav Goldberg, and Anders S{\o}gaard. 2016.
\newblock \href {https://doi.org/10.18653/v1/N16-1179} {Improving sentence
  compression by learning to predict gaze}.
\newblock In \emph{Proceedings of the 2016 Conference of the North {A}merican
  Chapter of the Association for Computational Linguistics: Human Language
  Technologies}, pages 1528--1533, San Diego, California. Association for
  Computational Linguistics.

\bibitem[{Mejia-Romero et~al.(2021)Mejia-Romero, Michaels, Eduardo~Lugo,
  Bernardin, and Faubert}]{mejia2021gaze}
Sergio Mejia-Romero, Jesse Michaels, J~Eduardo~Lugo, Delphine Bernardin, and
  Jocelyn Faubert. 2021.
\newblock Gaze movement’s entropy analysis to detect workload levels.
\newblock In \emph{Proceedings of International Conference on Trends in
  Computational and Cognitive Engineering: Proceedings of TCCE 2020}, pages
  147--154. Springer.

\bibitem[{Miller(2019)}]{Miller2019}
Tim Miller. 2019.
\newblock Explanation in artificial intelligence: Insights from the social
  sciences.
\newblock \emph{Artificial Intelligence}, 267:1--38.

\bibitem[{Morger et~al.(2022)Morger, Brandl, Beinborn, and
  Hollenstein}]{morger-etal-2022-cross}
Felix Morger, Stephanie Brandl, Lisa Beinborn, and Nora Hollenstein. 2022.
\newblock \href {https://aclanthology.org/2022.clasp-1.2} {A cross-lingual
  comparison of human and model relative word importance}.
\newblock In \emph{Proceedings of the 2022 CLASP Conference on
  (Dis)embodiment}, pages 11--23, Gothenburg, Sweden. Association for
  Computational Linguistics.

\bibitem[{Murrugarra-Llerena and Kovashka(2017)}]{murrugarra2017learning}
Nils Murrugarra-Llerena and Adriana Kovashka. 2017.
\newblock Learning attributes from human gaze.
\newblock In \emph{2017 IEEE Winter Conference on Applications of Computer
  Vision (WACV)}, pages 510--519. IEEE.

\bibitem[{Papoutsaki et~al.(2017)Papoutsaki, Laskey, and
  Huang}]{papoutsaki2017searchgazer}
Alexandra Papoutsaki, James Laskey, and Jeff Huang. 2017.
\newblock Searchgazer: Webcam eye tracking for remote studies of web search.
\newblock In \emph{Proceedings of the 2017 conference on conference human
  information interaction and retrieval}, pages 17--26.

\bibitem[{Parmar et~al.(2023)Parmar, Mishra, Geva, and
  Baral}]{parmar-etal-2023-dont}
Mihir Parmar, Swaroop Mishra, Mor Geva, and Chitta Baral. 2023.
\newblock \href {https://doi.org/10.18653/v1/2023.eacl-main.130} {Don{'}t blame
  the annotator: Bias already starts in the annotation instructions}.
\newblock In \emph{Proceedings of the 17th Conference of the European Chapter
  of the Association for Computational Linguistics}, pages 1779--1789,
  Dubrovnik, Croatia. Association for Computational Linguistics.

\bibitem[{Rajpurkar et~al.(2016)Rajpurkar, Zhang, Lopyrev, and
  Liang}]{rajpurkar-etal-2016-squad}
Pranav Rajpurkar, Jian Zhang, Konstantin Lopyrev, and Percy Liang. 2016.
\newblock \href {https://doi.org/10.18653/v1/D16-1264} {{SQ}u{AD}: 100,000+
  questions for machine comprehension of text}.
\newblock In \emph{Proceedings of the 2016 Conference on Empirical Methods in
  Natural Language Processing}, pages 2383--2392, Austin, Texas. Association
  for Computational Linguistics.

\bibitem[{Ribeiro et~al.(2023)Ribeiro, Brandl, S{\o}gaard, and
  Hollenstein}]{ribeiro2023webqamgaze}
Tiago Ribeiro, Stephanie Brandl, Anders S{\o}gaard, and Nora Hollenstein. 2023.
\newblock Webqamgaze: A multilingual webcam eye-tracking-while-reading dataset.
\newblock \emph{arXiv preprint arXiv:2303.17876}.

\bibitem[{Rosenfeld(2021)}]{10.5555/3463952.3463962}
Avi Rosenfeld. 2021.
\newblock Better metrics for evaluating explainable artificial intelligence.
\newblock In \emph{Proceedings of the 20th International Conference on
  Autonomous Agents and MultiAgent Systems}, AAMAS '21, page 45–50, Richland,
  SC. International Foundation for Autonomous Agents and Multiagent Systems.

\bibitem[{Rudin(2019)}]{rudin2019explaining}
Cynthia Rudin. 2019.
\newblock Stop explaining black box machine learning models for high stakes
  decisions and use interpretable models instead.
\newblock \emph{Nature Machine Intelligence}, 1(5):206--215.

\bibitem[{Samek et~al.(2021)Samek, Montavon, Lapuschkin, Anders, and
  Müller}]{XAIreview2021}
Wojciech Samek, Gr{\'e}goire Montavon, Sebastian Lapuschkin, Christopher~J.
  Anders, and Klaus-Robert Müller. 2021.
\newblock Explaining deep neural networks and beyond: A review of methods and
  applications.
\newblock \emph{Proceedings of the IEEE}, 109(3):247--278.

\bibitem[{Schmidt and Bie{\ss}mann(2019)}]{schmidt2019}
Philipp Schmidt and Felix Bie{\ss}mann. 2019.
\newblock Quantifying interpretability and trust in machine learning systems.
\newblock \emph{CoRR}, abs/1901.08558.

\bibitem[{Semmelmann and Weigelt(2018)}]{semmelmann_online_2018}
Kilian Semmelmann and Sarah Weigelt. 2018.
\newblock \href {https://doi.org/10.3758/s13428-017-0913-7} {Online
  webcam-based eye tracking in cognitive science: {A} first look}.
\newblock \emph{Behavior Research Methods}, 50(2):451--465.

\bibitem[{Serrano and Smith(2019)}]{serrano-smith-2019-attention}
Sofia Serrano and Noah~A. Smith. 2019.
\newblock \href {https://doi.org/10.18653/v1/P19-1282} {Is attention
  interpretable?}
\newblock In \emph{Proceedings of the 57th Annual Meeting of the Association
  for Computational Linguistics}, pages 2931--2951, Florence, Italy.
  Association for Computational Linguistics.

\bibitem[{Shrikumar et~al.(2017)Shrikumar, Greenside, and
  Kundaje}]{DBLP:journals/corr/ShrikumarGSK16}
Avanti Shrikumar, Peyton Greenside, and Anshul Kundaje. 2017.
\newblock Learning important features through propagating activation
  differences.
\newblock In \emph{Proceedings of the 34th International Conference on Machine
  Learning - Volume 70}, ICML'17, page 3145–3153.

\bibitem[{Sood et~al.(2020)Sood, Tannert, Frassinelli, Bulling, and
  Vu}]{sood-etal-2020-interpreting}
Ekta Sood, Simon Tannert, Diego Frassinelli, Andreas Bulling, and Ngoc~Thang
  Vu. 2020.
\newblock \href {https://doi.org/10.18653/v1/2020.conll-1.2} {Interpreting
  attention models with human visual attention in machine reading
  comprehension}.
\newblock In \emph{Proceedings of the 24th Conference on Computational Natural
  Language Learning}, pages 12--25, Online. Association for Computational
  Linguistics.

\bibitem[{Swartout and Moore(1993)}]{swartout}
William~R. Swartout and Johanna~D. Moore. 1993.
\newblock \emph{Explanation in Second Generation Expert Systems}, page
  543–585. Springer-Verlag, Berlin, Heidelberg.

\bibitem[{Thorn~Jakobsen et~al.(2023)Thorn~Jakobsen, Cabello, and
  S{\o}gaard}]{thorn-jakobsen-etal-2023-right}
Terne~Sasha Thorn~Jakobsen, Laura Cabello, and Anders S{\o}gaard. 2023.
\newblock \href {https://doi.org/10.18653/v1/2023.acl-long.59} {Being right for
  whose right reasons?}
\newblock In \emph{Proceedings of the 61st Annual Meeting of the Association
  for Computational Linguistics (Volume 1: Long Papers)}, pages 1033--1054,
  Toronto, Canada. Association for Computational Linguistics.

\bibitem[{Tokunaga et~al.(2013)Tokunaga, Iida, and
  Mitsuda}]{tokunaga2013annotation}
Takenobu Tokunaga, Ryu Iida, and Koh Mitsuda. 2013.
\newblock Annotation for annotation-toward eliciting implicit linguistic
  knowledge through annotation-(project note).
\newblock In \emph{Proceedings of the 9th Joint ISO-ACL SIGSEM Workshop on
  Interoperable Semantic Annotation}, pages 79--84.

\bibitem[{Voita et~al.(2019)Voita, Talbot, Moiseev, Sennrich, and
  Titov}]{voita-etal-2019-analyzing}
Elena Voita, David Talbot, Fedor Moiseev, Rico Sennrich, and Ivan Titov. 2019.
\newblock \href {https://doi.org/10.18653/v1/P19-1580} {Analyzing multi-head
  self-attention: Specialized heads do the heavy lifting, the rest can be
  pruned}.
\newblock In \emph{Proceedings of the 57th Annual Meeting of the Association
  for Computational Linguistics}, pages 5797--5808, Florence, Italy.
  Association for Computational Linguistics.

\bibitem[{Wallace et~al.(2019)Wallace, Tuyls, Wang, Subramanian, Gardner, and
  Singh}]{wallace-etal-2019-allennlp}
Eric Wallace, Jens Tuyls, Junlin Wang, Sanjay Subramanian, Matt Gardner, and
  Sameer Singh. 2019.
\newblock \href {https://doi.org/10.18653/v1/D19-3002} {{A}llen{NLP} interpret:
  A framework for explaining predictions of {NLP} models}.
\newblock In \emph{Proceedings of the 2019 Conference on Empirical Methods in
  Natural Language Processing and the 9th International Joint Conference on
  Natural Language Processing (EMNLP-IJCNLP): System Demonstrations}, pages
  7--12, Hong Kong, China. Association for Computational Linguistics.

\bibitem[{Wu et~al.(2020)Wu, Cha, Sulek, Zhou, Sundaram, Wachs, and
  Yu}]{wu2020eye}
Chuhao Wu, Jackie Cha, Jay Sulek, Tian Zhou, Chandru~P Sundaram, Juan Wachs,
  and Denny Yu. 2020.
\newblock Eye-tracking metrics predict perceived workload in robotic surgical
  skills training.
\newblock \emph{Human factors}, 62(8):1365--1386.

\bibitem[{Wu and Ong(2020)}]{wu-ong-2020-explain}
Zhengxuan Wu and Desmond~C. Ong. 2020.
\newblock \href {https://arxiv.org/abs/2101.00196} {On explaining your
  explanations of bert: An empirical study with sequence classification}.
\newblock \emph{arXiv preprint}.

\bibitem[{Xu et~al.(2015)Xu, Ehinger, Zhang, Finkelstein, Kulkarni, and
  Xiao}]{xu2015turkergaze}
Pingmei Xu, Krista~A Ehinger, Yinda Zhang, Adam Finkelstein, Sanjeev~R.
  Kulkarni, and Jianxiong Xiao. 2015.
\newblock \href {http://arxiv.org/abs/1504.06755} {Turkergaze: Crowdsourcing
  saliency with webcam based eye tracking}.

\bibitem[{Zaidan et~al.(2007)Zaidan, Eisner, and Piatko}]{zaidan2007using}
Omar Zaidan, Jason Eisner, and Christine Piatko. 2007.
\newblock Using “annotator rationales” to improve machine learning for text
  categorization.
\newblock In \emph{Human language technologies 2007: The conference of the
  North American chapter of the association for computational linguistics;
  proceedings of the main conference}, pages 260--267.

\bibitem[{Zhang et~al.(2019)Zhang, Chen, Xue, and
  Zhang}]{DBLP:journals/corr/abs-1911-09017}
Hao Zhang, Jiayi Chen, Haotian Xue, and Quanshi Zhang. 2019.
\newblock Towards a unified evaluation of explanation methods without ground
  truth.
\newblock \emph{CoRR}, abs/1911.09017.

\bibitem[{Zhang and Zhang(2019)}]{zhang-zhang-2019-using}
Yingyi Zhang and Chengzhi Zhang. 2019.
\newblock \href {https://doi.org/10.18653/v1/P19-1588} {Using human attention
  to extract keyphrase from microblog post}.
\newblock In \emph{Proceedings of the 57th Annual Meeting of the Association
  for Computational Linguistics}, pages 5867--5872, Florence, Italy.
  Association for Computational Linguistics.

\bibitem[{Zhou et~al.(2021)Zhou, Gandomi, Chen, and
  Holzinger}]{electronics10050593}
Jianlong Zhou, Amir~H. Gandomi, Fang Chen, and Andreas Holzinger. 2021.
\newblock Evaluating the quality of machine learning explanations: A survey on
  methods and metrics.
\newblock \emph{Electronics}, 10(5).

\end{thebibliography}

\label{lr:ref}
\bibliographystylelanguageresource{lrec-coling2024-natbib}
\bibliographylanguageresource{custom}

\end{document}